\documentclass[12pt]{article}

\usepackage[english]{babel}

\usepackage[a4paper,top=2.5cm,bottom=2.5cm,left=2.5cm,right=2.5cm,marginparwidth=1.75cm]{geometry}

\usepackage{graphicx}
\usepackage{subfig}

\usepackage{multirow}	
\usepackage{booktabs} 
\usepackage{makecell} % for multiline cells
\usepackage{amsmath}
\usepackage{graphicx}
\usepackage[colorlinks=true, allcolors=blue]{hyperref}
\usepackage{mathptmx}
\usepackage{newtxtext,newtxmath}

\usepackage{natbib}
\patchcmd{\bibfont}{\normalfont}{\normalfont\small}{}{} % 11 pt references font size

\usepackage{titlesec}

\titleformat{\section} % which section command to format
  {\fontsize{12}{14}\bfseries} % format for whole line
  {\thesection.} % how to show number (with dot)
  {1em} % space between number and text
  {} % formatting for just the text
  [] % formatting for after the text

\titleformat{\subsection} % which subsection command to format
  {\fontsize{12}{14}\bfseries} % format for whole line (12pt)
  {\thesubsection.} % how to show number (with dot)
  {1em} % space between number and text
  {} % formatting for just the text
  [] % formatting for after the text

\titlelabel{\thetitle.\quad} % Adding dot for section number

\newcommand{\aff}[1]{\par{#1}} % Removed \textit
\newcommand{\email}[1]{\par{#1}}

\makeatletter
\renewcommand{\@maketitle}{%
  \begin{center}%
    {\fontsize{14}{16}\bfseries\@title\par}%
    \vskip 1.5em%
    {\fontsize{10}{12}\@author}
    {\small\aff{$^{1}$Chair of Computational Modeling and Simulation}} 
    {\small\aff{Technical University of Munich, Munich, Germany}}
    {\small\aff{$^{2}$S.M.A.R.T. Construction Research Group, Division of Engineering}}
    {\small\aff{New York University Abu Dhabi (NYUAD), Saadiyat Island, United Arab Emirates}}
    {\email{\href{mailto:e.miguel.vega@tum.de}{miguel.vega@tum.de}}}
  \end{center}%
}
\makeatother

\title{BIMCaP: BIM-based AI-supported LiDAR-Camera Pose Refinement}

\author{{%
    M.A. Vega-Torres$^{1}$, 
    A. Ribic$^{1}$, 
    B. García de Soto$^{2}$ \& 
    A. Borrmann$^{1}$
}}

\renewenvironment{abstract}
 {\quotation\small\noindent\ignorespaces}
 {\endquotation}

\begin{document}
\maketitle

%The Abstract must include only these six short sentences: 
% 1. What is the problem being studied?
% 2. What is the research question to be answered?
% 3. What is the methodology used to answer the research question? 
% 4. What is the answer to the research question obtained with the use of the selected method?
% 5. What is the meaning of the answer, and why (and for whom) is the answer important?
% 6. How does the obtained answer inspire future research and development?

\begin{abstract}
\textbf{Abstract.}

This paper introduces BIMCaP, a novel method to integrate mobile 3D sparse LiDAR data and camera measurements with pre-existing building information models (BIMs), enhancing fast and accurate indoor mapping with affordable sensors. 
BIMCaP refines sensor poses by leveraging a 3D BIM and employing a bundle adjustment technique to align real-world measurements with the model.
Experiments using real-world open-access data show that BIMCaP achieves superior accuracy, reducing translational error by over 4 cm compared to current state-of-the-art methods. 
This advancement enhances the accuracy and cost-effectiveness of 3D mapping methodologies like SLAM. 
BIMCaP's improvements benefit various fields, including construction site management and emergency response, by providing up-to-date, aligned digital maps for better decision-making and productivity. Link to the repository: \url{https://github.com/MigVega/BIMCaP}.

\end{abstract}
% Define custom link colors -> after making the title, and writing the abstract only remove the colors of the references
\definecolor{linkcolor}{rgb}{0,0,0} % Black color for links
\definecolor{citecolor}{rgb}{0,0,0} % Black color for citation links
\definecolor{urlcolor}{rgb}{0,0,0} % Black color for URLs
\hypersetup{
    % colorlinks=true,
    linkcolor=linkcolor,
    citecolor=citecolor,
    urlcolor=urlcolor
}

\section{Introduction}
%   (1/5) Background or Context:  also in the following point
%   (2/5) Significance or Importance (Motivation):
This research explores the convergence of 3D Building Information Modeling (BIM) with real-world 3D reconstruction, utilizing cost-effective RGB and LiDAR sensors. Traditional 3D data acquisition methods, such as terrestrial laser scanning, present challenges in terms of high costs and time-intensive procedures, particularly in the context of construction site monitoring or disaster relief. The practical implementation of faster 3D data acquisition methods has been made possible by the advancement of simultaneous localization and mapping (SLAM). However, state-of-the-art algorithms still encounter challenges to accurately map complex and dynamic environments, such as construction sites. % \citep{hiltiChallenge2022}. % removed citation due to space

BIM has become a revolutionary technology within the Architecture, Engineering, and Construction (AEC) domain, offering comprehensive geometric and semantic information throughout a building's life cycle. In this context, BIMs provide a valuable foundation for rectifying data acquired through SLAM algorithms in real time using low-cost sensors. 
The automatic alignment of RGB and LiDAR data with the BIM holds significant potential to rapidly create precise 3D maps in GPS-denied environments (such as indoors). 
This alignment not only facilitates safety monitoring and quality management but also contributes to the quick development of a digital twin, providing an accurate 3D representation of actual asset states.

%   (3/5) Research Question or Problem Statement:

In this research, we address several critical questions central to advancing state-of-the-art technologies in the field of sensor pose correction for accurate 3D reconstruction. 
By utilizing the geometric and semantic information inherent in a BIM, we examine the potential of a bundle adjustment module to refine drifted sensor poses. 
Additionally, we explore methodologies for effectively combining calibrated sparse LiDAR data with RGB images to produce detailed depth maps, aiming for a comprehensive reconstruction. 
Furthermore, we evaluate semantic segmentation algorithms in complex indoor construction settings and develop a strategy for improving their performance. 
We demonstrated the improvement in performance through extensive experiments on the publicly available ConSLAM dataset \citep{trzeciak2023conslamExtension}. 
In this research, we aim to contribute to the development of robust and efficient techniques for 3D reconstruction, which has implications for a range of applications, including construction site management, emergency response, and beyond.

%   (4/5) Scope (assumptions) and Limitations: 
As we delve into the intricacies of our methodology, it is essential to acknowledge the scope and assumptions of our method. 
Since our goal is to have a robust method using sensors with a reduced field of view (FoV), such as solid-state LiDARs or RGB-D cameras, we only use the LiDAR information in the FoV of the camera, ignoring the rest of the available points. 
In our approach, we also assume that there is an initial rough alignment of the drifted trajectory with the reference map.

%   (5/5) Overview of the Paper: - Since this is very standard, I could remove it if space is needed. 
This paper is structured as follows: Section \ref{sec_related_work} offers an in-depth exploration of related research efforts. 
Section \ref{sec_methodology} delineates the proposed framework split into three main steps. 
Subsequently, Section \ref{sec_results} presents the findings derived from numerous experiments conducted on real-world construction site data. 
Finally, Section \ref{sec_conclusions} provides the culmination of this work, offering conclusive insights and avenues for future research.

% % % % % % % % % % % % % % % % % % % % % % % % % % % % % % % % %
\section{Related Work}

\label{sec_related_work}
Several studies have approached the alignment of RGB images with BIMs in two main ways: (1) as a global localization problem and (2) as a pose-tracking problem. 

In the global localization problem, \cite{Acharya.2022} introduced BIM-PoseNet, utilizing synthetic images from a 3D indoor model to achieve a 2-meter accurate camera pose without an initial position. 
%\cite{Haque.2020} localized a unmanned aerial vehicle in the BIM coordinate system by detecting doors and windows in RGB images, using You Only Look Once (YOLO) for object detection and ORB-SLAM2 for 3D mapping. 

In the pose-tracking approach, \cite{Kropp.2018} focused on image-to-4D BIM registration using line segments as features, with manual intervention for initial registration. 
\cite{Boniardi.05.03.2019} proposed a clutter-handling method using a convolutional neural network for layout prediction and a particle filter algorithm for pose tracking using a floor plan as a reference map.
The method proposed by \cite{Dantas:2022:FBI} aims at quickly correcting camera poses using vanishing points, lines, and synthetic renders created from a BIM.

Other methods addressed the challenge of creating a coherent 3D map of the environment aligned with a given reference map.
\cite{vega:2023:BIM_SLAM} used a BIM to align and correct 360-degree LiDAR measurements, which initial poses were calculated with a LiDAR-based SLAM algorithm.
\cite{Sokolova.2022} presented the Floorplan-Aware Camera Poses Refinement (FACaP) method, aligning Visual-SLAM maps with floor plans using semantic segmentation and an optimization model considering geometric, floor-to-plane and wall-to-floorplan terms for map correction.

However, most of these methods were tested in indoor residential apartments without the level of clutter, dynamic elements, and changing lighting conditions present in real-world construction sites.
Furthermore, the literature review shows accuracy metrics for diverse methods; however, all were evaluated on separate, unrelated case studies. 
This fact highlights the necessity for a standardized dataset (i.e., BIM and synchronized sensor information) tailored explicitly to real-world construction site environments, enabling fair and consistent ranking.

\section{Methodology}
\label{sec_methodology}

\begin{figure}[!htb]
    \centering
	\includegraphics[width=\textwidth]{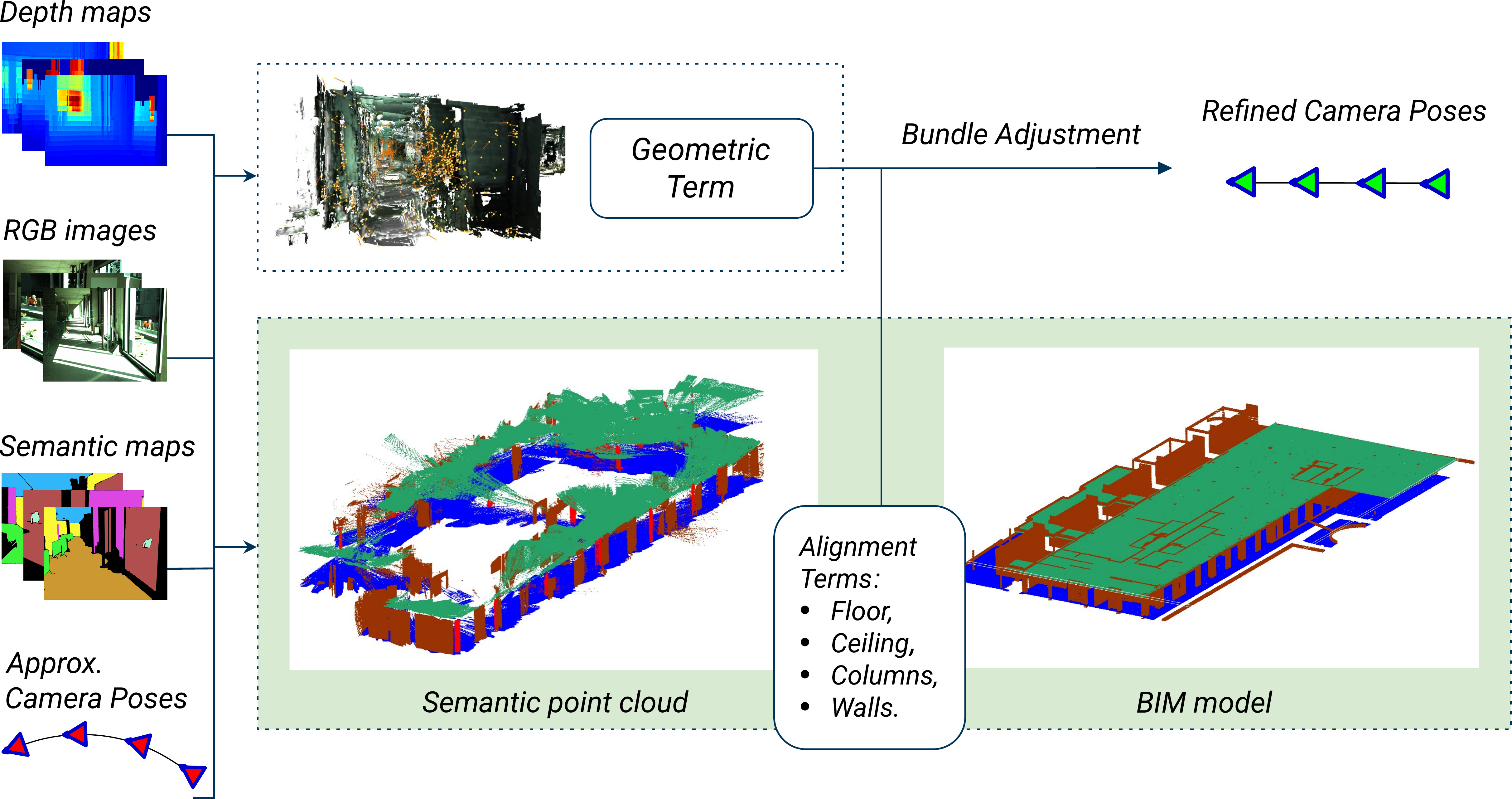} 
    \caption{Overview of the proposed BIMCaP framework for sensor pose refinement. The depth maps are used to project the semantic maps (created from the images) into the 3D space using the approximated initial poses (drifted due to SLAM). Different terms aim to correlate the data measured over permanent elements (i.e., reliable landmarks) with the BIM. Moreover, a geometric term ensures geometric consistency among real-world images.
    }
    \label{fig:overview}
\end{figure}

We propose a framework designed to align a sequence of synchronized LiDAR scans and RGB images with a 3D BIM, thereby refining the initial approximated camera poses, which inherently suffer from drift owing to the characteristics of SLAM algorithms. Our framework can be divided into three significant steps.
\textbf{Step 1.} The initial step of our methodology involves fusing camera images and sparse LiDAR scans in precise depth maps. This process is facilitated through a hybrid approach employing interpolation and a deep learning (DL) technique, which then allows the projection of the pixel information (such as semantic information) into the 3D space.
\textbf{Step 2.} Subsequently, in the second step, semantic segmentation is applied to the images, enabling the detection of permanent elements such as walls, columns, and floors within the reconstructed 3D map. Simultaneously, a point cloud and a vectorized floor plan with semantic information are created from the BIM. This vectorized semantic floor plan will be used as a reference map for the alignment of the real-world data.
\textbf{Step 3.} In the third step, we employ a statistical approach to generate initial synthetic camera poses. These poses are then refined through a bundle adjustment (BA) module, which integrates custom cost functions. These functions are designed to iteratively enhance the accuracy of sensor poses, thereby ensuring optimal alignment between the generated map and the semantically vectorized floor plan from the BIM. This refinement process selectively considers only permanent elements, which are identified through semantic segmentation in real-world images and projected into three-dimensional space using the previously estimated depth maps.
Fig. \ref{fig:overview} illustrates the proposed semantic-aware pose optimization framework.

\subsection{Step 1: LiDAR and camera fusion}
% Overview
To fuse the information from the LiDAR and the camera, we first project the visible point cloud (in the FoV of the camera) into the image, and then we aim to generate a dense map that is coherent with the image and LiDAR information.
% LiDAR -  Camera
The projection of the LiDAR points to the camera image is made with the intrinsic and extrinsic parameters of the camera and with the package provided by \cite{trzeciak2023conslamExtension}; this package ensures the camera image is undistorted, and only the corresponding (timestamped synchronized) LiDAR points of the small FoV of the camera are projected from the 3D space to the 2D image. 
Upon this step, we now have depth information for several pixels of the image. This depth is, however, very sparse since we are working with a 360$^{\circ}$ LiDAR. 
It is essential to mention that to ensure that the method works appropriately with sensors with a reduced FoV (such as solid-state LiDARs or RGB-D cameras), we only use the LiDAR information in the FoV of the camera. 

% Depth Completion
The sparsity of the point cloud would not be sufficient to leverage all the information from the image in the 3D space; therefore, we subsequently aim to create a dense depth map using the point cloud and the corresponding camera image.

\begin{figure}[!htb]
    \centering
    \subfloat[]{\label{fig:originalImage}{
	\includegraphics[height=2.8cm]{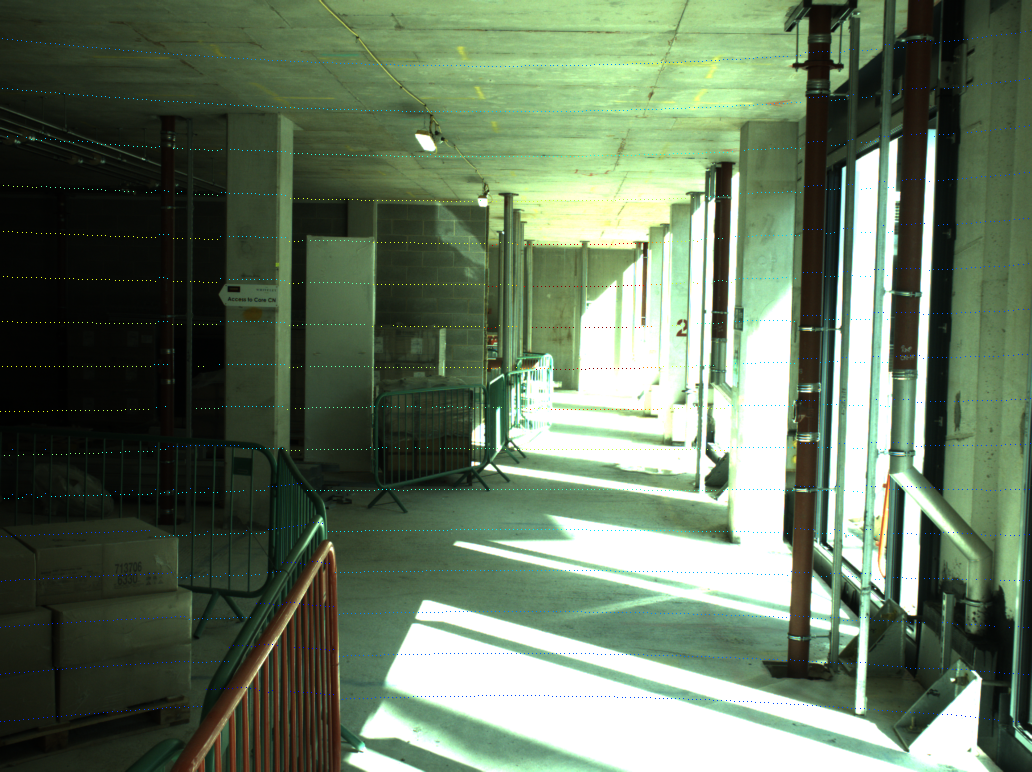}}}\hfill % for2columns width=0.48\textwidth
 	\subfloat[]{\label{fig:depth0}{
	\includegraphics[height=2.8cm]{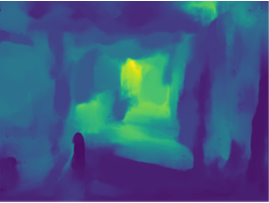}}}\hfill
 	\subfloat[]{\label{fig:depth1}{
	\includegraphics[height=2.8cm]{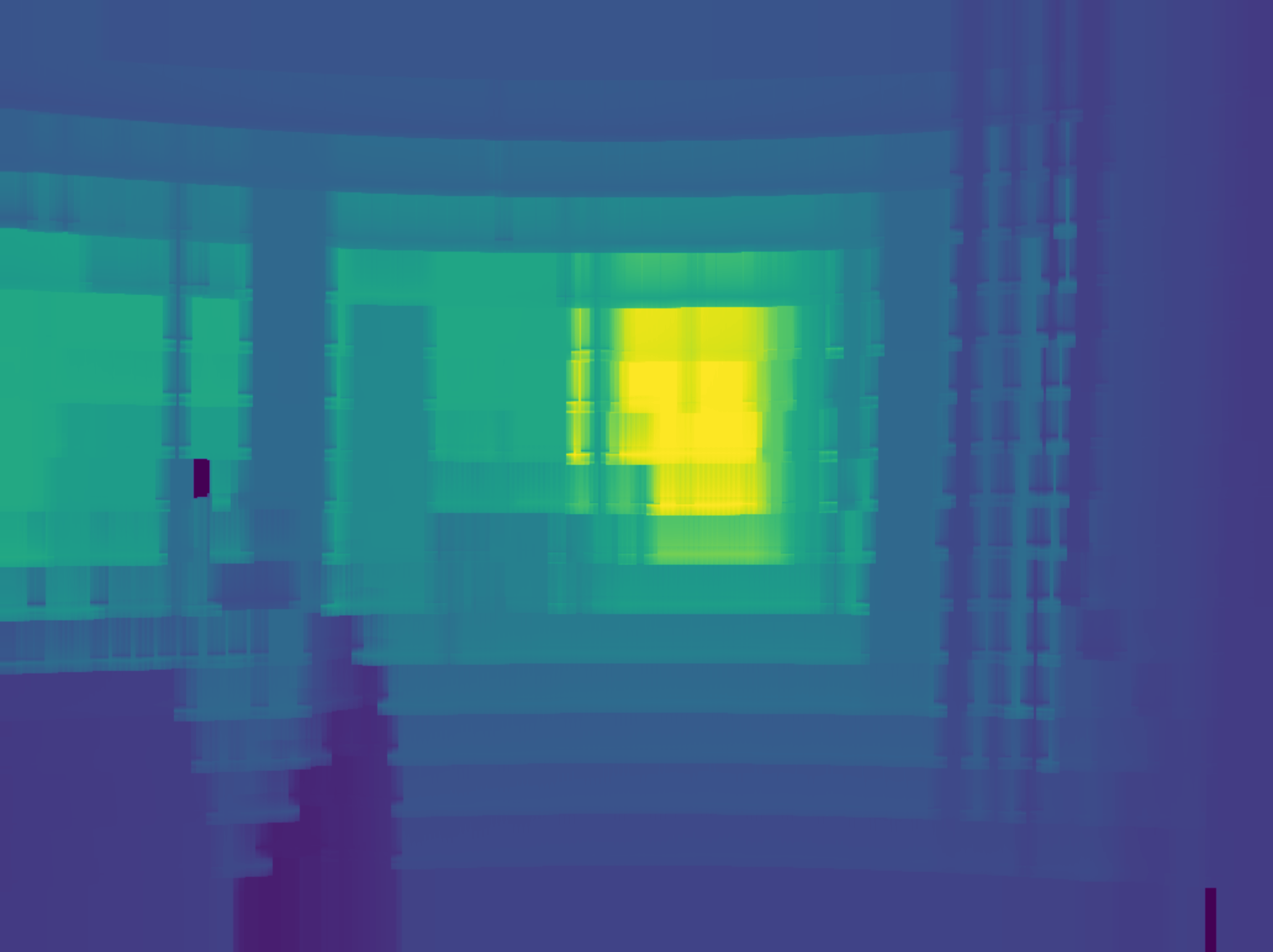}}}\hfill
  	\subfloat[]{\label{fig:depth2}{
	\includegraphics[height=2.8cm]{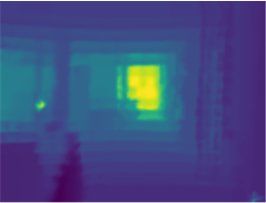}}}
    \caption{Depth completion with sparse LiDAR point cloud: (a) original image from the ConSLAM dataset with original sparse projected LiDAR scan; (b) depth map using only CompletionFormer; (c) depth map using only linear interpolation and (d) using linear interpolation and CompletionFormer. It is evident that (d) yields the best results since it is smother than (c) and more coherent with the measurements than (b).}
    \label{fig:depthmaps}
\end{figure}

Currently, numerous DL methods serve for depth estimation, yet many of them are optimized for outdoor environments (such as the KITTI dataset). Therefore, their accuracy tends to decline in indoor settings, which constitutes the focus of our investigation. 
Following extensive experimentation with various methodologies, we chose to adopt a hybrid method that combines linear interpolation with CompletionFormer (CF) \citep{zhang2023completionformer}. 
Fig. \ref{fig:depthmaps} illustrates the results of this hybrid approach, showing the original CF output alongside the refined outcome involving an initial linear interpolation.

% \nointerlineskip
\nointerlineskip

\subsection{Step 2: Semantically enriched maps} 
In this step, we aim to create maps that will allow the sensor pose correction in the subsequent step. This step is divided into two sub-steps: Firstly, we create a reference semantic vectorized floor plan from the BIM, and secondly, we enrich the 3D map created with real-world data with semantic information. 
This semantic enrichment serves a pivotal role in distinguishing permanent elements within real-world data, such as walls, columns, and floors, which can be reliably aligned with the BIM.

\subsubsection{Reference map}
To prepare for implementing the pose correction module, we simplify the 3D BIM into a 2D semantic vectorized floor plan. Since walls and columns are perpendicular to the XY plane, this reduction not only retains all vertical structural element information but also allows efficient pose optimization in subsequent stages. 

\begin{figure}[!htb]
    \centering
	\subfloat[]{\label{fig:bim}{
	\includegraphics[height=3cm]{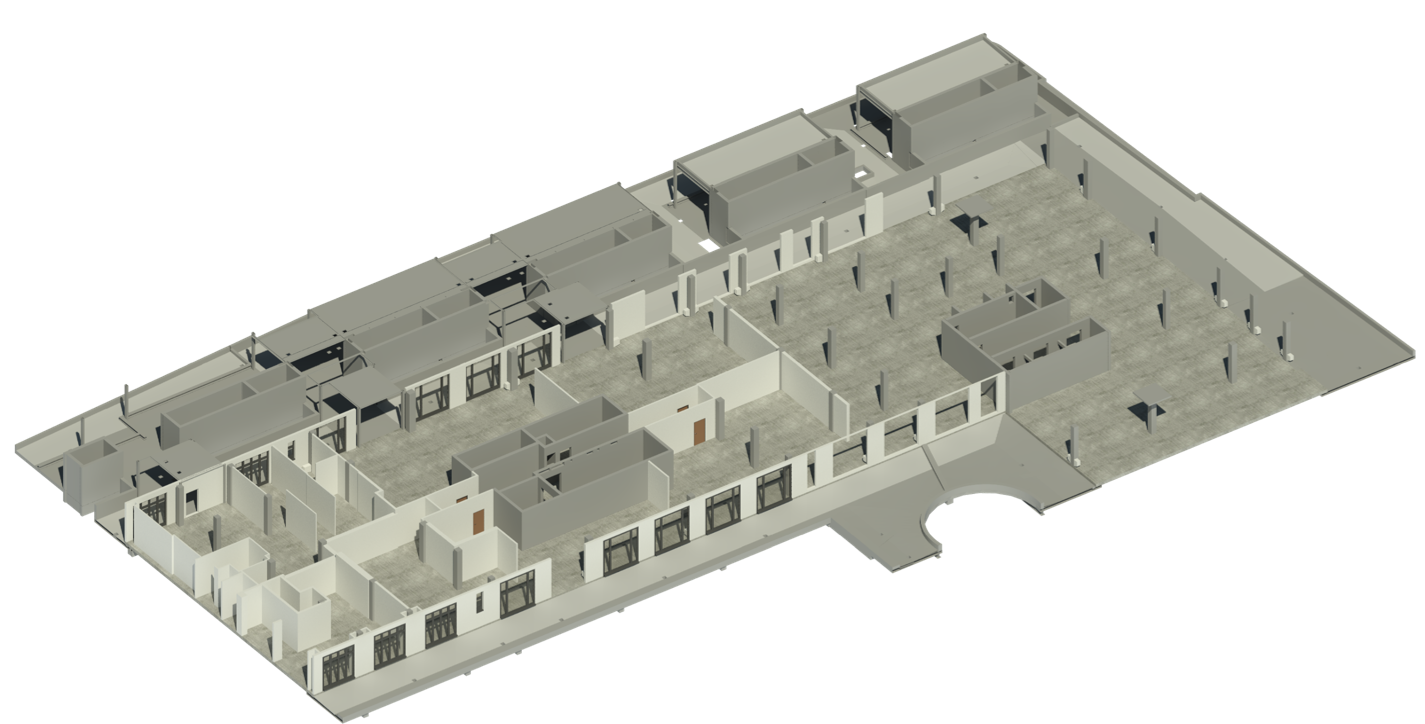}}}\hfill 
 	\subfloat[]{\label{fig:pc_bim}{
	\includegraphics[height=3cm]{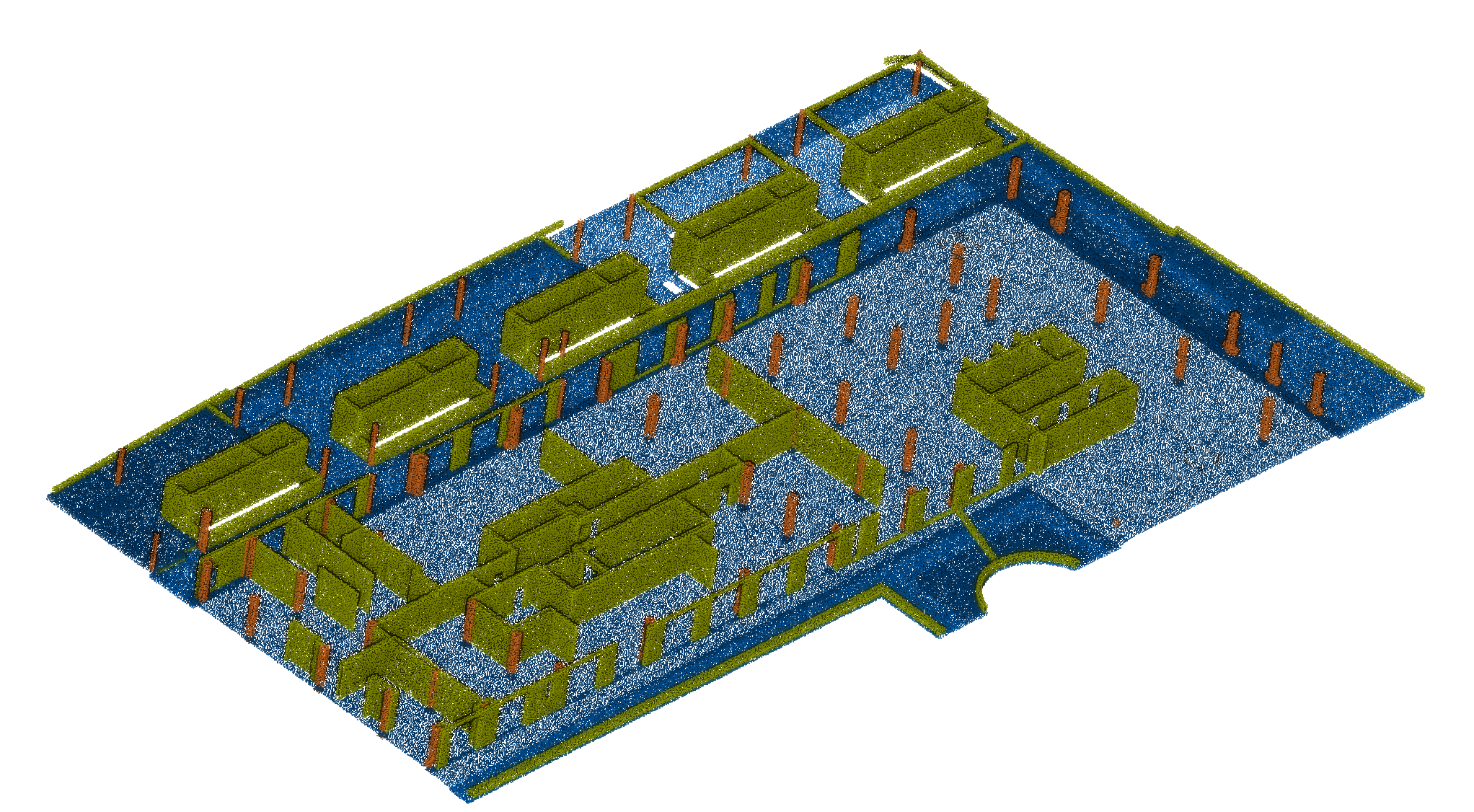}}}\hfill
 	\subfloat[]{\label{fig:vec_fp}{
	\includegraphics[height=3cm]{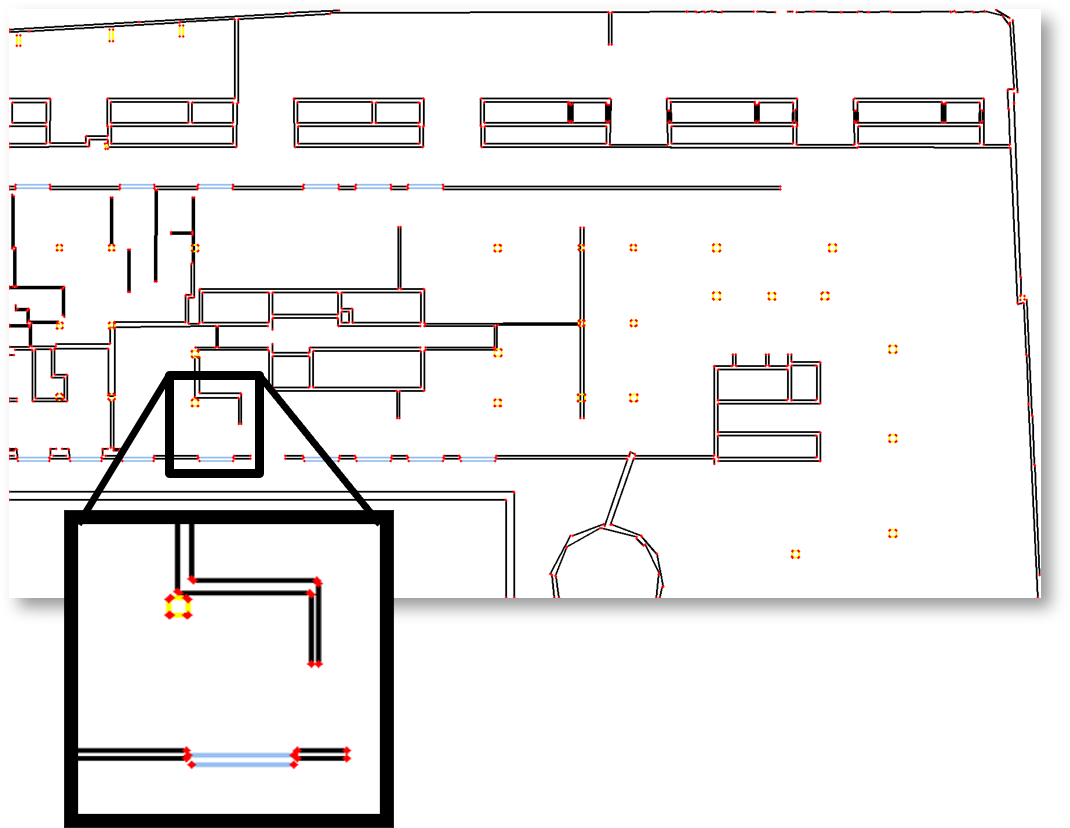}}}
    \caption{Reference map preparation: (a) original 3D BIM (without ceiling); (b) uniformly sampled 3D point cloud with semantic information from the BIM; and (c) vectorized semantic floor plan, from which the walls and columns (in black and yellow) are used for pose refinement in the subsequent pose optimization step.}
    \label{fig:ref_map}
\end{figure}

To generate the 2D semantic vectorized floor plan, the BIM undergoes conversion from Industry Foundation Classes (IFC) format to OBJ format using ifcConvert. 
Following this, distinct OBJ files are generated for each entity within the model (e.g., walls, columns, floor, ceiling, windows, and doors). Then, uniform point cloud sampling is applied to each OBJ file, and the resulting semantically enriched synthetic point clouds are merged into a single one. 
An illustration of such a point cloud can be observed in Fig. \ref{fig:pc_bim}.

The created synthetic 3D point cloud is projected vertically into 2D images within a specified height range, typically within  $\pm$ 20 cm from the floor level. 
Semantic labels are utilized to filter each element in the point cloud. 
Subsequently, image processing methods such as contour and line detection are employed to identify line segments representing individual elements in the 2D projection. 
These detected lines are then consolidated, including their start and end points, to form the vectorized semantic floor plan. 
A resulting floor plan is depicted in Fig. \ref{fig:vec_fp}.

\subsubsection{Semantic segmentation of real-world data}
\label{step2b}
\begin{figure}[!htb]
    \centering
	\subfloat[]{\label{fig:seg_a}{
	\includegraphics[width=0.23\textwidth]{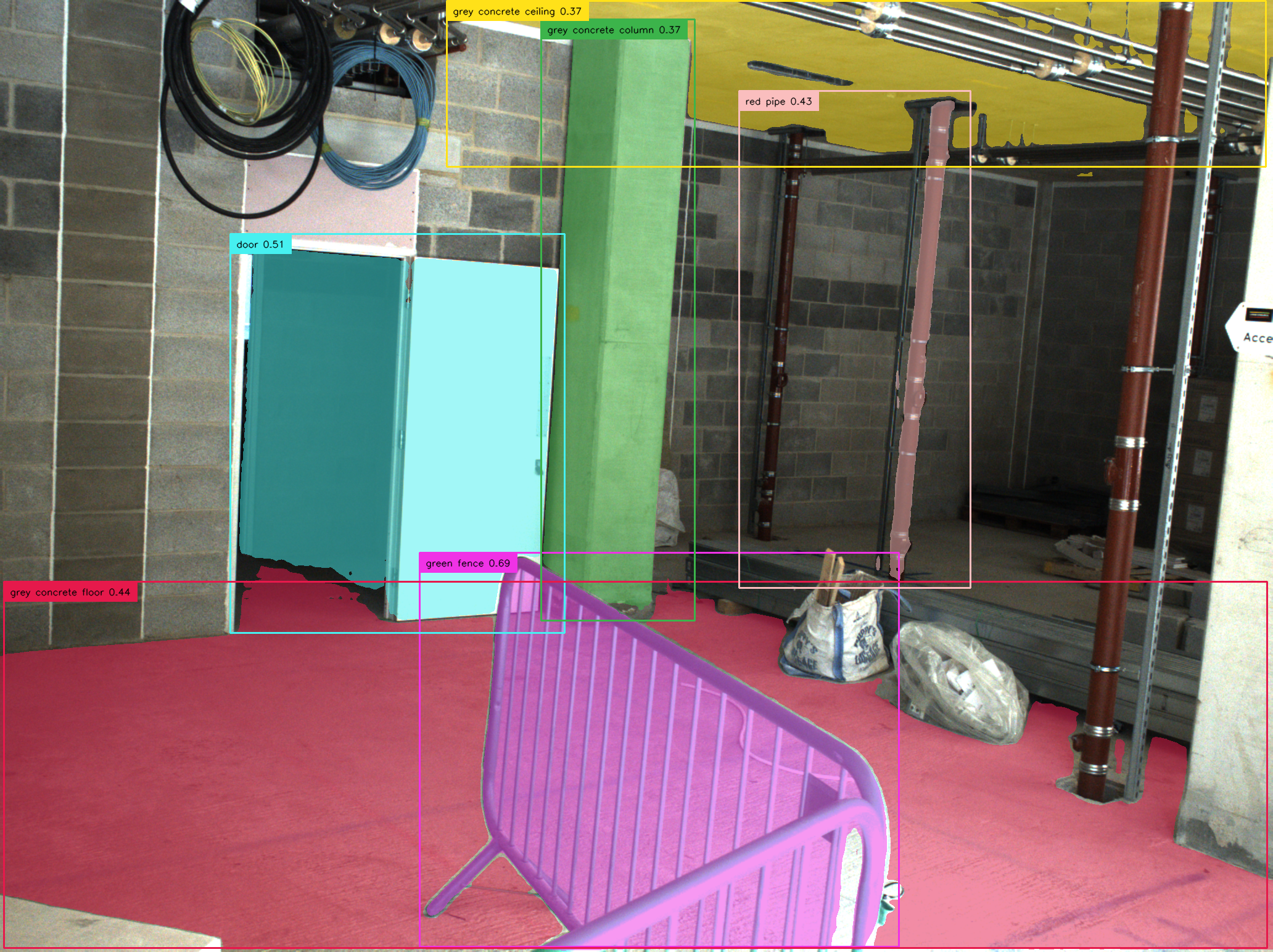}}}\hfill 
 	\subfloat[]{\label{fig:seg_b}{
	\includegraphics[width=0.23\textwidth]{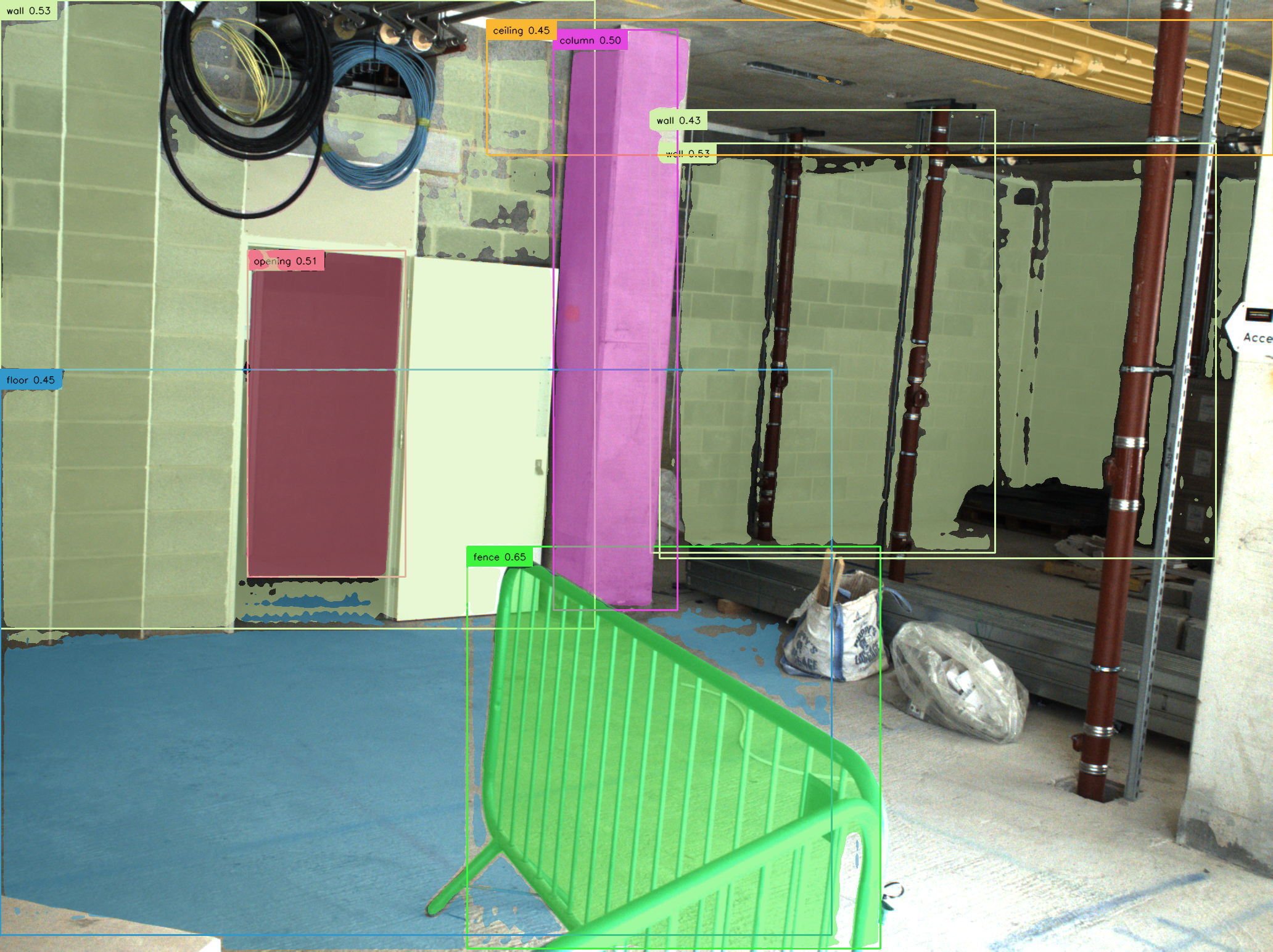}}}\hspace{0.7cm}
 	\subfloat[]{\label{fig:seg2_a}{
	\includegraphics[width=0.23\textwidth]{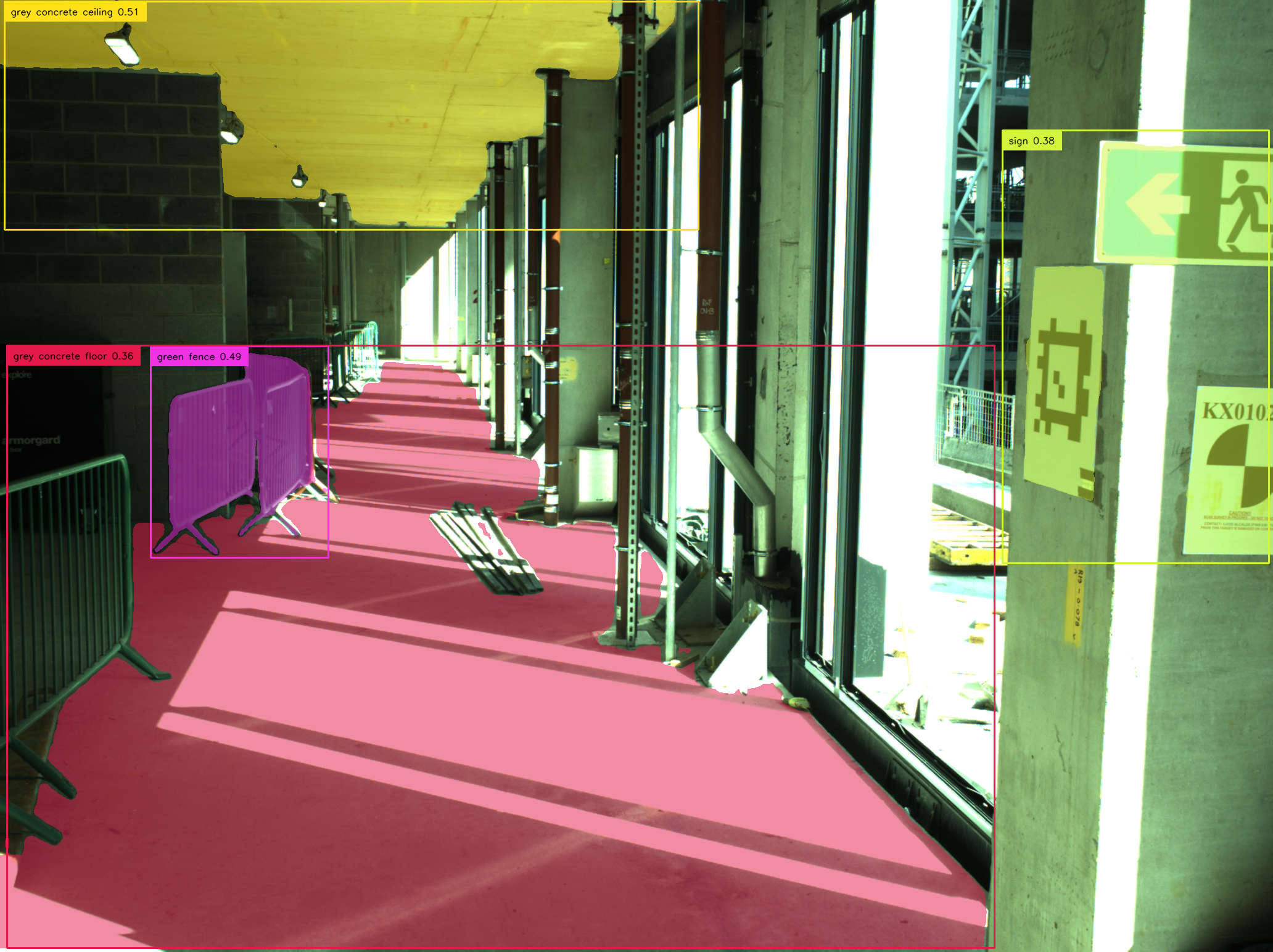}}}\hfill
 	\subfloat[]{\label{fig:seg2_b}{
	\includegraphics[width=0.23\textwidth]{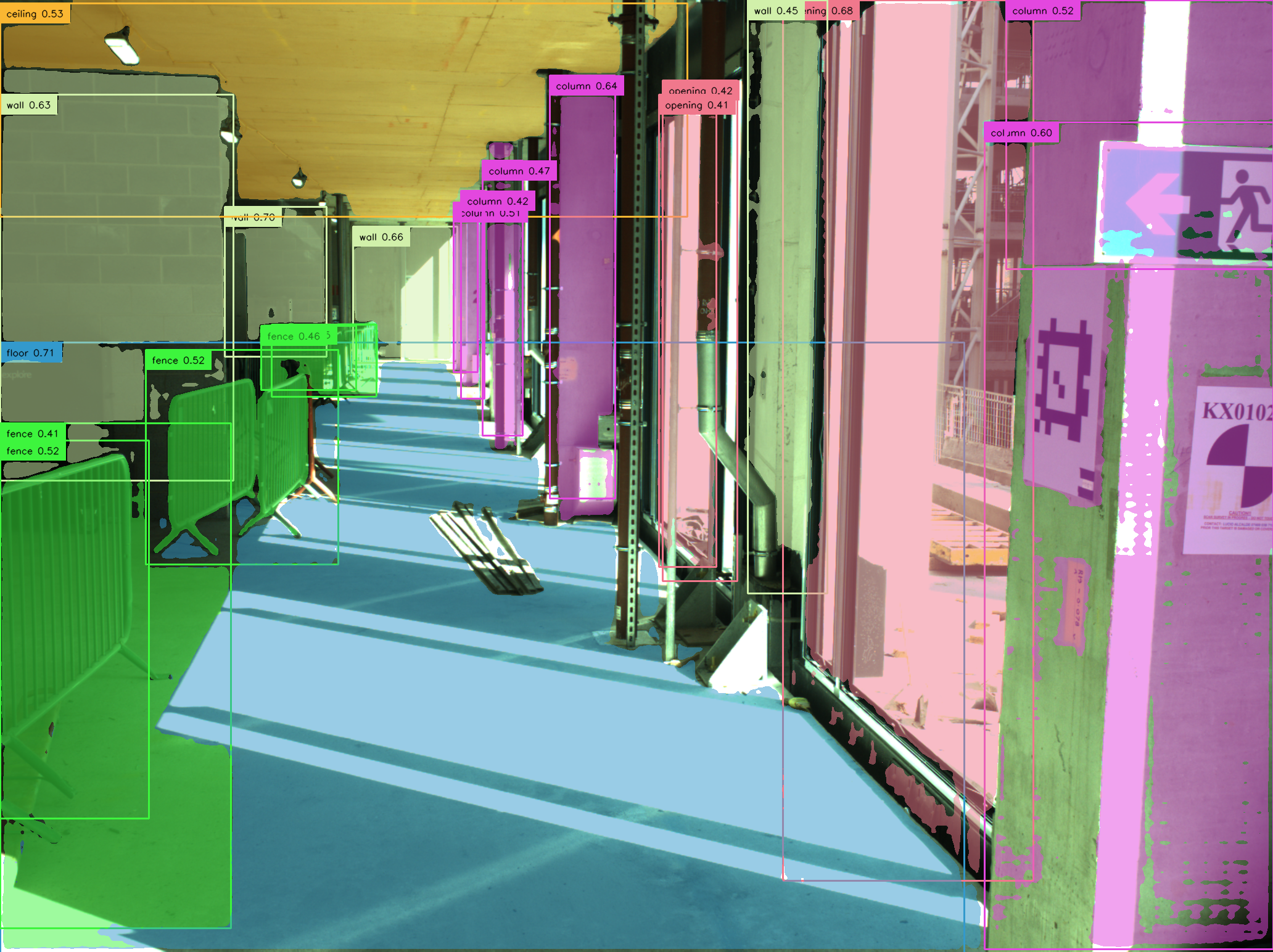}}}
    \caption{Semantic segmentation over 2D images of the ConSLAM dataset: (a) inference with the original Grounding DINO algorithm (b) inference result after replacing DINO with pre-trained RTMDet for object detection. (b) comprehends predicted labels for the walls in the background, which are critical for our camera pose refinement framework. Similarly, (c) and (d) are, respectively, the results of Grounding DINO and the results of our proposed pipeline.}
    \label{fig:seg_result}
\end{figure}

% Image semantic segmentation 
To filter permanent elements that we can match from the real-world data with the BIM, we leverage state-of-the-art image semantic segmentation algorithms. 
More specifically, we use a modified version of Grounding DINO \citep{liu2023grounding}. 
However, for the object detection task, we replace the DINO algorithm with a tiny version of the RTMDet algorithm \citep{lyu2022rtmdet} pre-trained with the COCO dataset and 250 labeled images of the ConSLAM dataset, which contains custom classes typical of a construction site. 
These images were labeled semi-automatically using the Computer Vision Annotation Tool (CVAT).
Thus, our approach enables the detection of objects of interest, expanding beyond the foreground elements identified by the original Grounding DINO version.
Fig. \ref{fig:seg_result} illustrates the results of the semantic enrichment before and after the proposed enhancement, and Fig. \ref{fig:gt_map_segmented} shows the top view of the resulting semantically enriched 3D point cloud after projecting the semantic labels to the 3D space with the previously generated depth maps. 
In this last figure, it is also visible that we can now filter walls, columns, floor, and ceiling points in the depth maps, which can reliably be used for registration with the BIM and, therefore, for camera pose optimization.
It is worth mentioning that the floor and ceiling predicted labels were also used to optimize the depth maps, creating smoother surfaces in these regions with blurring operations in the 2D depth maps.

\nointerlineskip

%%%%%%%%%%%%%%%%%%%%%%%%%%%%%%%%%%%%%%%%%%%%%%%%%%%%%%5
\subsection{Step 3: Sensor pose calculation and refinement}
The initial approximations of sensor poses are ideally determined using a Visual-SLAM framework. 
However, our experimentation with cutting-edge SLAM algorithms such as DROID-SLAM or Go-SLAM yielded unsatisfactory results when applied to the ConSLAM dataset, functioning correctly only for limited segments of the trajectory. 
Despite these limitations, advancements in odometry systems suggest that addressing this challenge will become feasible in the future. 
Therefore, and since we aim to refine slightly drifted poses and experiment under different magnitudes of drift, we opted to create a synthetic trajectory instead, simulating the output of a SLAM framework. 
This process is explained in the following subsection. 
Subsequently, we introduce the method that is used to improve the accuracy of these poses.

\subsubsection{Synthetic pose calculation}
\label{step3a}
To make sure our approach matches the usual trajectory patterns seen in existing SLAM systems and to have the flexibility to study how stable our method is when it comes to convergence with different starting positions, we carefully engineer synthetic trajectories. 
When creating these trajectories, we focus on replicating the gradual drifting feature of SLAM-generated trajectories. In other words, we want the error at each pose to slowly increase over time.

Therefore, we model the translation offset from the original sensor pose \( \Delta T_ {i+1} \) as a normally distributed random variable with mean \( \Delta t_ {i}\) and variance  \( \sigma_t^2 \).  
Formally, $
\Delta T_{i+1} \sim \mathcal{N}\left(\Delta t_i, \sigma_t^2\right) $ with $\Delta T_1 \sim \mathcal{N}\left(0, \sigma_t^2\right)$ where \( \Delta t_ {i} \) is the previously sampled offset value, and the variance \( \sigma_t^2 \) is an adjustable hyper-parameter which would determine the offset of the trajectory from the ground truth poses.
Regarding the camera rotation, we randomly sample degree offsets $\Delta \phi\sim \mathcal{N}\left(0, \sigma_p^2\right), \Delta \theta \sim$ $\mathcal{N}\left(0, \sigma_{t h}^2\right)$ around pitch and yaw directions. Fig. \ref{fig:synthetic} presents the resulting map using a synthetically drifted trajectory.

\subsubsection{Pose optimization}
\begin{figure}[!htb]
    \centering
	\subfloat[]{\label{fig:gt_map_segmented}{
	\includegraphics[height=3.2cm]{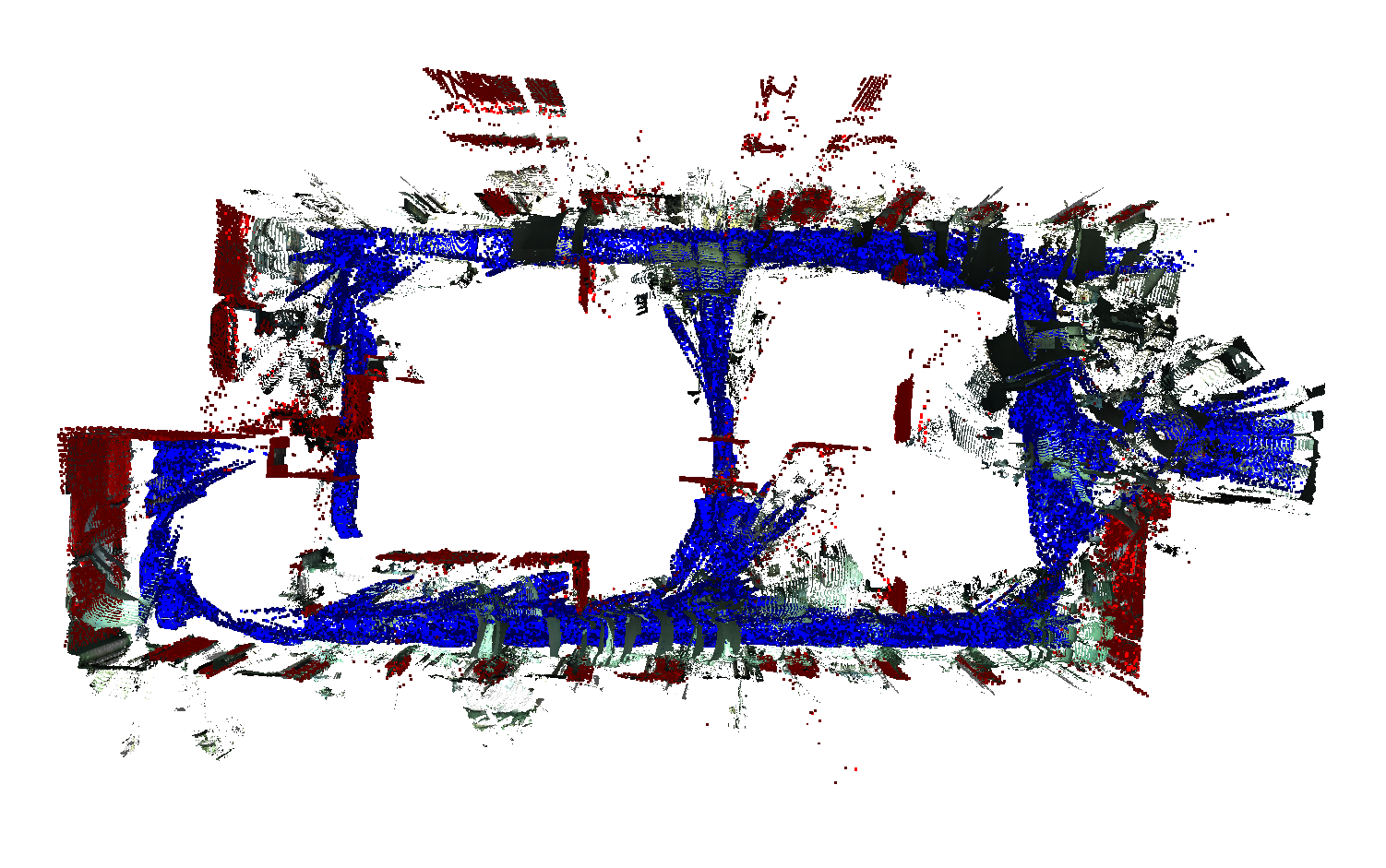}}}\hfill 
 	\subfloat[]{\label{fig:synthetic}{
	\includegraphics[height=3.2cm]{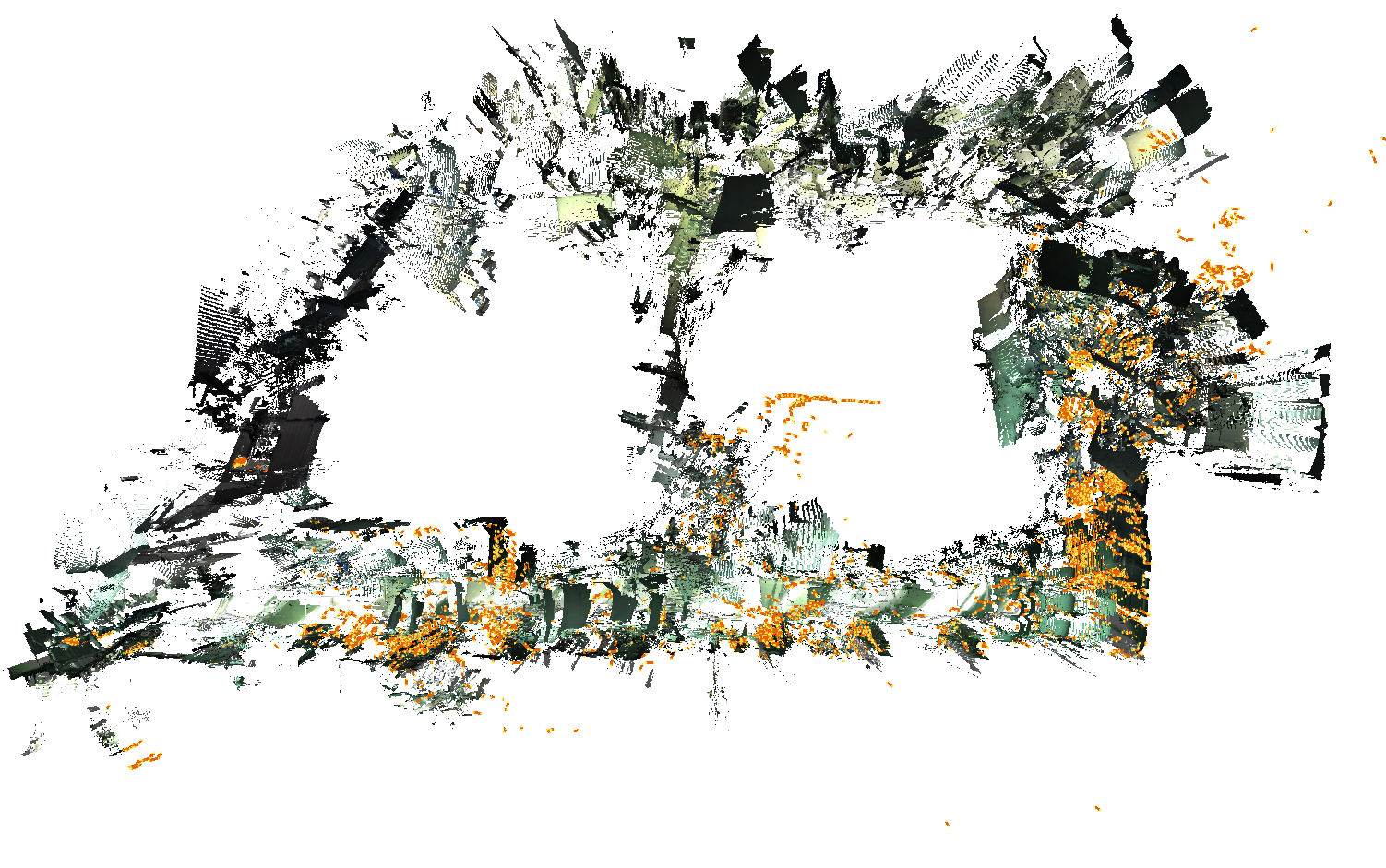}}}\hfill
 	\subfloat[]{\label{fig:keypoints}{
	\includegraphics[height=2.8cm]{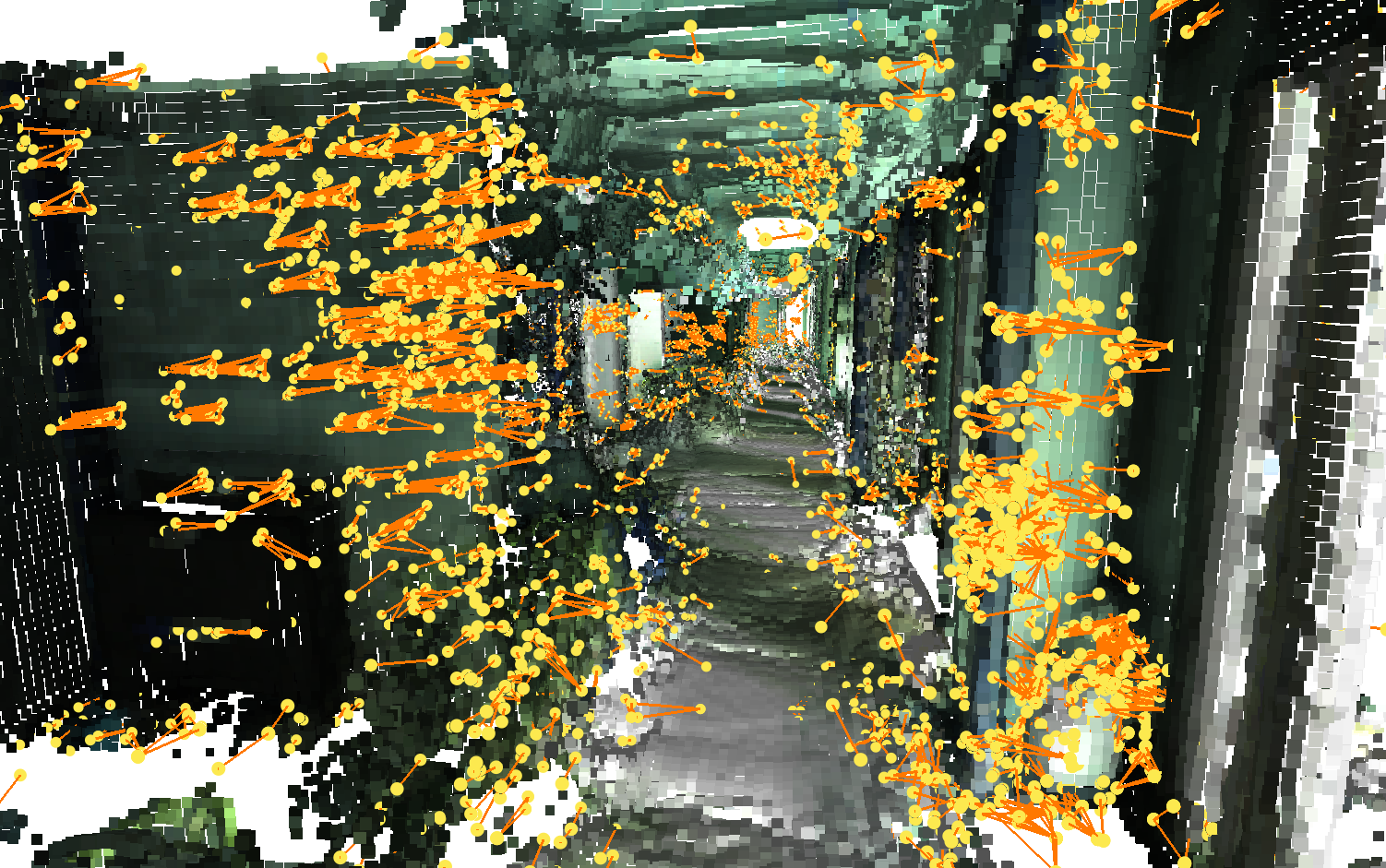}}}
    \caption{Features used for optimization (a) Top view semantic segmented map generated with the ground truth poses and the segmentation results of walls (in red) and floor (in blue) as explained in Section \ref{step2b}; (b) map created with synthetic poses of Exp. 1 (obtained as explained in Section \ref{step3a}), here the COLMAP features are visible; (c) view from an indoor observer's perspective of the point cloud with highlighted Scale-Invariant Feature Transform (SIFT) features used in the geometric term for optimization.}
    \label{fig:several_result}
\end{figure}

Inspired by the FaCAP framework \citep{Sokolova.2022}, we consider several terms in our cost function to achieve sensor pose refinement with the BIM using BA. 
To create a consistent 3D map from the real-world sequential images, we use a geometric term that encapsulates the divergence between 3D point estimations from two distinct viewpoints. 
In other words, the geometric term considers photogrammetric constraints, and in our case, we use COLMAP to obtain features and correspondences among sequential images. Fig. \ref{fig:synthetic} and \ref{fig:keypoints} visualize some of these features. 

Furthermore, we use a floor term designed to ensure that segmented points corresponding to the floor lie within a single plane and close to the floor surface defined in the BIM. 
In a parallel manner, we introduce a ceiling term that incorporates information from the model's ceiling for optimization purposes. 
In addition, we include wall and column terms. These elements are crucial for correcting rotations around the vertical axis (i.e., yaw variations) and horizontal translations. 
% \nointerlineskip

%%%%%%%%%%%%%%%%%%%%%%%%%%%%%%%%%%%%%%%%%%%%%%%%%%%%%%%%%%%%%%%5
\section{Experiments and results}
\label{sec_results}

\subsection{Dataset and evaluation details}
To ensure reproducibility and enable benchmarking, we tested the developed method on the ConSLAM dataset \citep{trzeciak2023conslamExtension}. 
ConSLAM represents a pioneering effort, offering the first open-access dataset acquired in an indoor cluttered construction site. 
This dataset encompasses sequences of RGB and LiDAR data together with Terrestrial Laser Scanning (TLS) point clouds. The latter was leveraged as a resource for the generation of a BIM with centimeter-level accuracy. 
The GT poses of ConSLAM were calculated using SLAM2REF \citep{SLAM2REF:vega2024:paper}, an enhanced version of BIM-SLAM \citep{vega:2023:BIM_SLAM} and OGM2PGBM \citep{vega:2022:2DLidarLocalization} for large-scale maps, which is robust to LiDAR motion distortion and Scan-Map deviations.

To quantify the quality of the whole trajectory before and after pose optimization, we used the standardized root mean square error (RMSE) of the absolute trajectory error (ATE) in position (also referred to as translation) and in rotation. 
Moreover, for better comparison, we incorporated some of the metrics introduced in \citep{Sokolova.2022} including the Map Mean Entropy (MME), Mean Plane Variance (MPV), and the Nearest Neighbor Distance (NND). 
The MME serves to assess the quality of 3D maps, with a higher MME signifying favorable alignment between the input cloud and the reference map. 
The MPV evaluates the variance among planes within the map, with lower MPV values indicating more uniform and well-defined surfaces. The NND quantifies the average distance between adjacent points in the point cloud, with smaller NND values indicating denser point clouds. 
The MME value of 0.761, calculated using the GT poses (as shown in \ref{tab:merged_table}), represents the optimal alignment between the real-world point cloud and the BIM model. For understanding, this value would be zero if no deviations between the actual environment and the model (Scan-BIM deviations) exist.
\subsection{Pose refinement results}
The results of our framework are compared against the state-of-the-art FaCAP pipeline \citep{Sokolova.2022} and evaluated meticulously with three different experiments. 
The first experiment consists of a synthetic trajectory that has an offset of around 1.4 meters in translation and 10 degrees in rotation (Exp. 1), the second one has an offset of only 30.3 cm in translation and 8.82 deg in rotation  (Exp. 2), and the third one only has rotation offset of 9.6 degrees (Exp. 3).

Table \ref{tab:merged_table} shows initial metrics based on ground truth and synthetic poses for Exp. 1, along with results after pose optimization using various terms of the FACaP pipeline and the proposed BIMCaP framework.

\begin{table}[!tbp]
    \centering
    \caption{Comparison of validation measurements for Exp. 1 using the different methods. All values are in meters except for the rotational ATE, which is given in degrees. The best overall results are highlighted in bold, while the best results per method are underlined. G, F, W, Co, and Ce stand for the geometric, floor, wall, column, and ceiling terms, respectively.}
    \begin{tabular}{lcccccccccc}
    \toprule
    \textbf{Source/Method} & \thead{G} & \thead{F} & \thead{W} & \thead{Co} & \thead{Ce} & \thead{MME$\downarrow$} & \thead{MPV$\downarrow$} & \thead{NND$\downarrow$} & \thead{ATE$_\text{pos}$$\downarrow$} & \thead{ATE$_\text{rot}$$\downarrow$}\\
     % & & & & & & & & & & \\
    \midrule
    GT poses & - & - & - & - & - & 0.761 & 0.040 & 0 & 0 & 0 \\
    \midrule
    Exp. 1 & - & - & - & - & - & 1.027 & 0.059 & 0.557 & 1.391 & 9.99 \\
    \midrule
    FACaP & \checkmark & \checkmark & \checkmark & - & - & 0.979 & 0.054 & \underline{0.503} & \underline{1.321} & 15.40 \\
      & \checkmark & - & - & - & - & 1.013 & 0.058 & 0.566 & 1.385 & \underline{\textbf{8.84}} \\
     & - & \checkmark & - & - & - & \underline{0.966} & \underline{0.053} & \underline{0.503} & 1.358 & 16.50 \\
    & - & - & \checkmark & - & - & 1.031 & 0.059 & 0.545 & 1.378 & 9.82 \\
    \midrule
   BIMCaP & \checkmark & \checkmark & \checkmark & \checkmark & \checkmark & \underline{\textbf{0.956}} & \underline{\textbf{0.052}} & 0.460 & \underline{\textbf{1.281}} & 11.84 \\
     & \checkmark & \checkmark & \checkmark & - & \checkmark & 0.959 & 0.052 & \underline{\textbf{0.456}} & \underline{\textbf{1.281}} & 11.81 \\
      & \checkmark & \checkmark & \checkmark & - & - & 0.975 & 0.053 & 0.505 & 1.311 & 12.58 \\
      & - & \checkmark & - & - & - & 0.966 & 0.054 & 0.519 & 1.351 & 13.63 \\
      & - & - & \checkmark & \checkmark & - & 1.034 & 0.059 & 0.549 & 1.378 & \underline{9.75} \\
       & - & - & - & - & \checkmark & 0.982 & 0.054 & 0.480 & 1.387 & 12.71 \\
    \bottomrule
    \end{tabular}
    \label{tab:merged_table}
\end{table}
% \nointerlineskip

%%%%%%%%%%%%%%%%%%%%%%%%%%%%%%%%%%%%%%%%%%%%%%v
% Discussion -> Analysis
The findings from Exp. 1 (Tab. \ref{tab:merged_table}) emphasize the efficacy of utilizing all terms for optimizing translational errors. 
However, this approach may not consistently yield optimal results when addressing rotational errors. 
Notably, while BIMCaP demonstrates a superior reduction in translational error by 4 cm compared to FACaP, both methodologies become trapped in a local minimum, impeding the accurate optimization of the poses. 
This issue can be attributed to the substantial difference between the synthetic poses and the ground truth.

\begin{table}[!htb]
    \centering
    \caption{Pose optimization results for Exp. 2 and 3: given a small translational and rotational offset. In addition to the ATE$_\text{pos}$ and ATE$_\text{rot}$, we provide the RMSE for the Yaw, Pitch, and Roll axes separately in degrees.}
    \begin{tabular}{lccccc}
    \toprule
    \textbf{Source/Method} & \thead{ATE$_\text{pos}$ (cm)$\downarrow$} & \thead{ATE$_\text{rot}$ (deg)$\downarrow$}& \textbf{Yaw}$\downarrow$& \textbf{Pitch}$\downarrow$& \textbf{Roll}$\downarrow$\\
    \midrule
    Exp. 2  before optim. & 30.3& 8.82& 4.31& 4.31& 0.29\\
    \hline
    FACaP \citep{Sokolova.2022}&  \textbf{30.2}&  7.70& 3.79& 3.79& \textbf{0.21}\\
    BIMCaP & 30.4&   \textbf{5.61}& \textbf{2.73}& \textbf{2.75} & 0.23\\
    \midrule
    Exp. 3  before optim.& 0 & 9.60& 4.70 & 4.72 & 0.29\\
    \hline
    FACaP \citep{Sokolova.2022}&  \textbf{6.2}& 7.92& 3.91& 3.90& \textbf{0.19} \\
    BIMCaP & 7.2&  \textbf{6.04} & \textbf{2.96}& \textbf{2.96} & 0.22\\
    \bottomrule
    \end{tabular}
   \label{tab:exp2_3}
\end{table}

\begin{figure}[!tb]
    \centering
	\subfloat[]{\label{fig:translationDev_rot}{
	\includegraphics[width=0.48\textwidth]{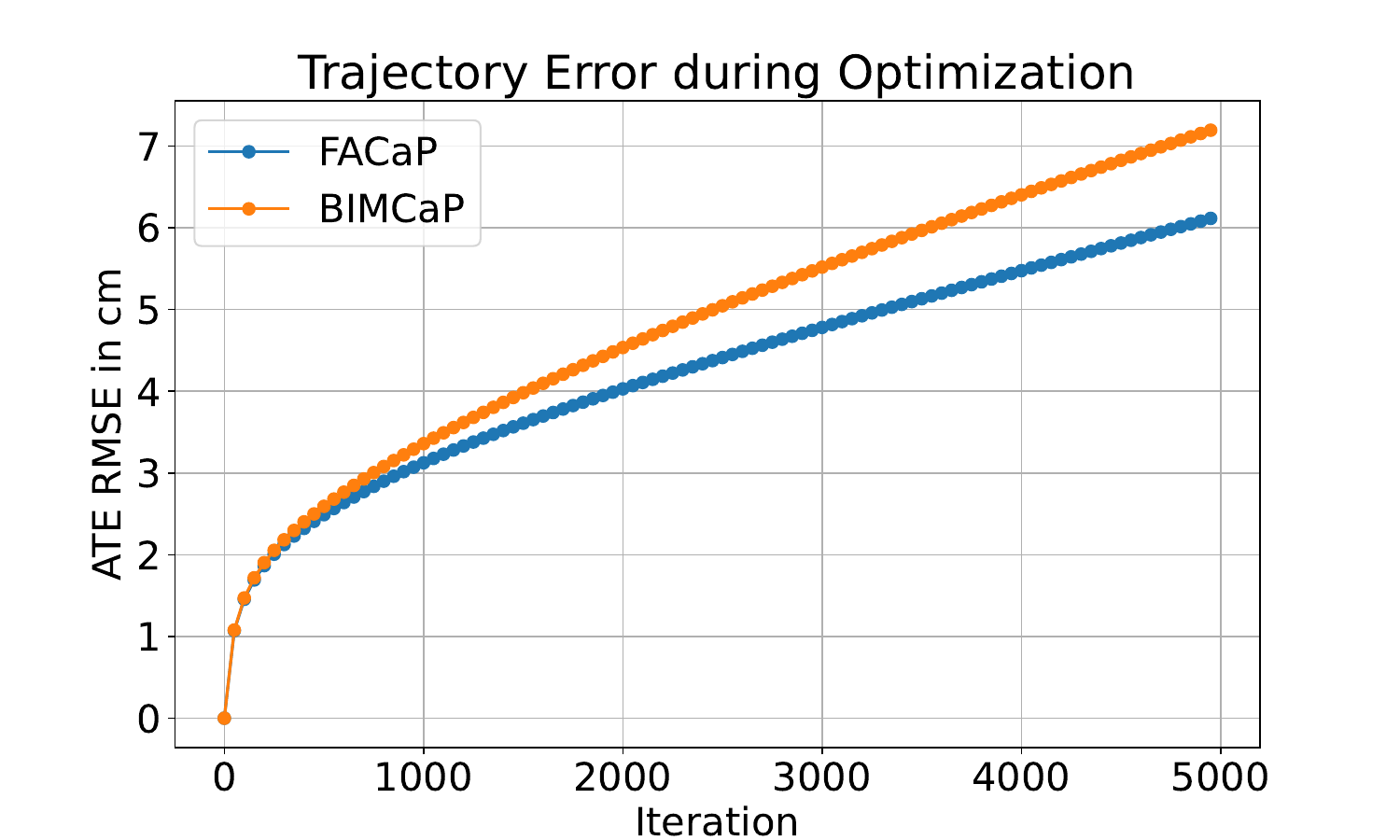}}}\hfill 
 	\subfloat[]{\label{fig:rotDev_rot}{
	\includegraphics[width=0.48\textwidth]{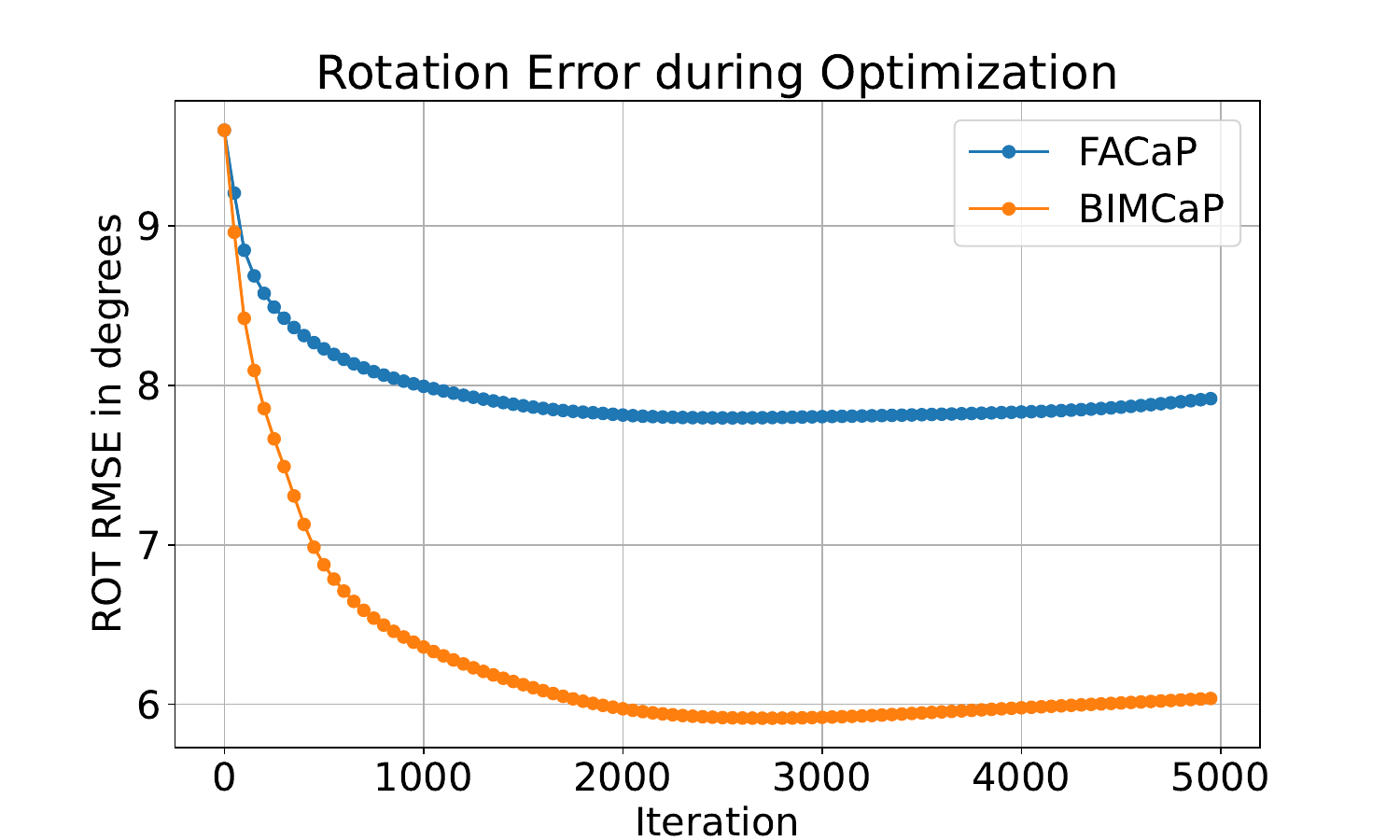}}}
    \caption{Development of the translational (a) and rotational error (b) given only a rotational offset as described for Exp. 3.
    }
    \label{fig:rot_offset}
\end{figure}

\begin{figure}[!htb]
    \centering
	\subfloat[]{\label{fig:side_a}{
	\includegraphics[width=0.41\textwidth]{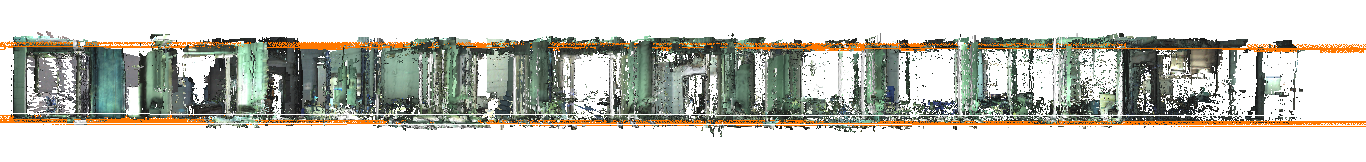}}}\hfill % for2columns width=0.48\textwidth
 	\subfloat[]{\label{fig:side_b}{
	\includegraphics[width=0.41\textwidth]{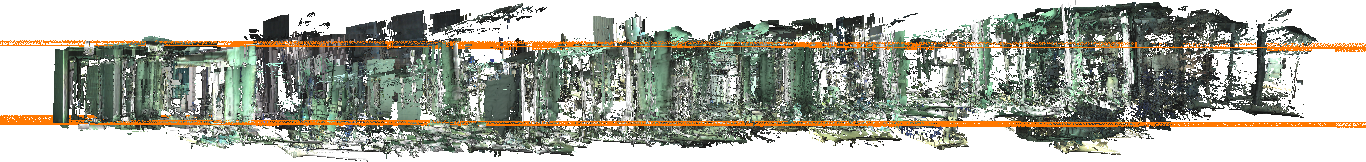}}}\hspace{0cm}
 	\subfloat[]{\label{fig:side_c}{
	\includegraphics[width=0.41\textwidth]{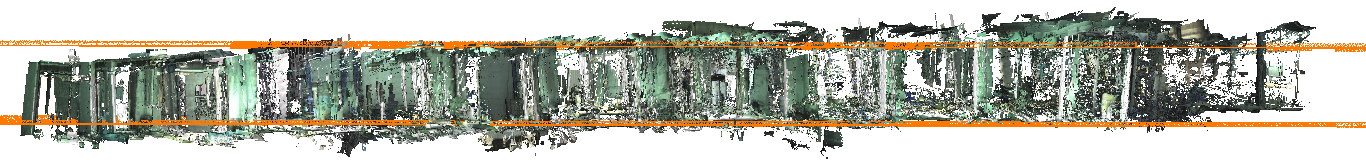}}}\hfill % for2columns width=0.48\texwidth
 	\subfloat[]{\label{fig:side_d}{
	\includegraphics[width=0.41\textwidth]{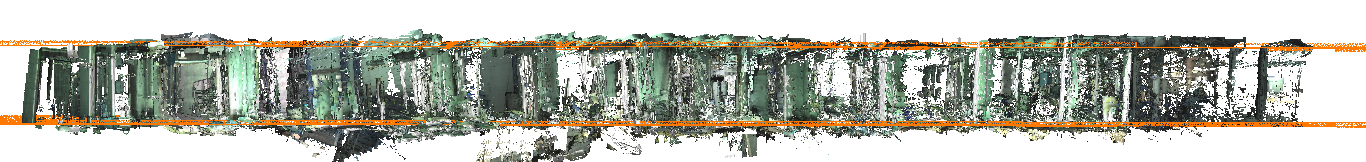}}}
    \caption{Side views of the different maps. (a) ground truth map; (b) map created with synthetic poses of Exp. 1; (c) map after FACaP optimization and (d) after BIMCaP optimization. The BIMCaP result shows better alignment with the real floor and ceiling planes.} 
    \label{fig:sideview_result}
\end{figure}

Exp. 2 and 3 results (Table \ref{tab:exp2_3} and Fig. \ref{fig:rot_offset}) indicate superior performance of both methods in optimizing rotational errors over translational errors. Notably, BIMCaP significantly enhances yaw and pitch angles during the pose optimization process.
Fig. \ref{fig:sideview_result} illustrates how BIMCaP aligns the floor and ceiling points to the correct planes, contrary to FACaP, which tries to fit a plane among the given measurements without any reference.

% \nointerlineskip

 % \nointerlineskip

% limitations 
Exp. 3 exposes a limitation in our approach, as optimizing trajectory with only rotational offsets resulted in unintended translations.
This could be due to the simultaneous optimization of both translation and rotation, causing discrepancies. Additionally, the challenge of accurately calculating sensor poses is intensified by the reduced FoV and the sparse ground truth poses.

\section{Conclusion and future directions}
\label{sec_conclusions}
We have demonstrated that BIMCaP enables alignment and correction of a sequence of the camera and reduced FoV LiDAR measurements with a BIM; the technique takes into consideration only reliable selected semantic landmarks (in our case, floor, walls, columns, and ceiling) for the drift correction.
Moreover, we evaluated our technique in the open-access ConSLAM dataset and compared it against a state-of-the-art method, ensuring reproducibility and benchmarking.  
In future work, we aim to improve the accuracy of the optimization process and add global registration and change detection capabilities to our framework.

 \nointerlineskip

\section{Acknowledgement}
This research is part of the INTREPID project funded by EU's Horizon 2020 program  (Grant agreement ID: 883345) and the Research Unit 5672 funded by the German Research Foundation (DFG) (Grant ID: 517965147). This work has also benefited from the collaboration with the NYUAD Center for Interacting Urban Networks (CITIES), funded by Tamkeen under the NYUAD Research Institute Award CG001.
Shaowen Qi's contributions to data labeling and to the semantic segmentation step are also acknowledged.

\bibliographystyle{agsm} %agsm
% \bibliographystyle{abbrvnat} % style that supports DOIs

% Adjusting the font size locally
{\small 
\bibliography{references}
}
\end{document}